\documentclass{article}%
\usepackage{graphicx}
\usepackage{amsmath}
\usepackage{amsfonts}
\usepackage{amssymb}%
\pdfoutput=1
\newtheorem{theorem}{Theorem}

\newtheorem{example}[theorem]{Example}

\newtheorem{remark}[theorem]{Remark}

\def\baselinestretch{1.0}
\ifx\pdfoutput\relax\let\pdfoutput=\undefined\fi
\newcount\msipdfoutput
\ifx\pdfoutput\undefined\else
\ifcase\pdfoutput\else
\ifx\paperwidth\undefined\else
\ifdim\paperheight=0pt\relax\else\pdfpageheight\paperheight\fi
\ifdim\paperwidth=0pt\relax\else\pdfpagewidth\paperwidth\fi
\fi\fi\fi
\begin{document}

\title{Geometric Analysis of the Conformal Camera for Intermediate-Level Vision and
Perisaccadic Perception}
\author{Jacek Turski\\University of Houston-Downtown\\Department of Computer and Mathematical Sciences}
\date{V2 with figures\\
March 2009}
\maketitle

\begin{abstract}
A binocular system developed by the author in terms of projective Fourier
transform (PFT) of the conformal camera, which numerically integrates the
head, eyes, and visual cortex, is used to process visual information during
saccadic eye movements. Although we make three saccades per second at the
eyeball's maximum speed of $700$ deg/sec, our visual system accounts for these
incisive eye movements to produce a stable percept of the world. This visual
constancy is maintained by neuronal receptive field shifts in various
retinotopically organized cortical areas prior to saccade onset, giving the
brain access to visual information from the saccade's target before the eyes'
arrival. It integrates visual information acquisition across saccades. Our
modeling utilizes basic properties of PFT. First, PFT is computable by FFT in
complex logarithmic coordinates that approximate the retinotopy. Second, a
translation in retinotopic (logarithmic) coordinates, modeled by the shift
property of the Fourier transform, remaps the presaccadic scene into a
postsaccadic reference frame. It also accounts for the perisaccadic
mislocalization observed by human subjects in laboratory experiments. Because
our modeling involves cross-disciplinary areas of conformal geometry, abstract
and computational harmonic analysis, computational vision, and visual
neuroscience, we include the corresponding background material and elucidate
how these different areas interwove in our modeling of primate perception. In
particular, we present the physiological and behavioral facts underlying the
neural processes related to our modeling. We also emphasize the conformal
camera's geometry and discuss how it is uniquely useful in the
intermediate-level vision computational aspects of natural scene understanding.


\end{abstract}

\section{Introduction}

In the last few years, we have developed projective Fourier analysis for
computational vision in the framework of the representation theory of the
semisimple Lie group $\mathbf{SL}(2,\mathbb{C})$
\cite{Turski1,Turski2,Turski3,Turski4,Turski5}. It was done by restricting the
group representations to the image plane of the conformal camera---the camera
with image projective transformations given by the action of $\mathbf{SL}%
(2,\mathbb{C})$. This analysis provides an efficient image representation and
processing that are not only well adapted to the projective transformations of
retinal images, but are also to the retinotopic mappings of the brain's
oculomotor and visual pathways. This latter assertion stems from the fact that
the projective Fourier transform (PFT) is computable by a fast Fourier
transform algorithm (FFT) in coordinates given by a complex logarithm that
transforms PFT into the standard Fourier integral and at the same time
approximates the retinotopic mappings \cite{Schwartz}.

However, the conformal camera is somewhat abstract and noticeably different
than any other camera model used in computer vision. Nevertheless, its
remarkable advantages are revealed to us every time we model specific
physiological processes involved in visual perception. For instance, one could
reasonably expect that a stationary camera and a moving object is similar to a
moving camera and a stationary because the relative position of the camera and
the object could be the same in both cases. Remarkably, it fails in primate
vision systems. In fact, when the image of a fast-moving object sweeps across
a static retina, though we are normally aware of its motion, we fail to detect
the comparable motion of images as they sweep across the retina during fast
eye movements. Computational modeling presented in this article demonstrates
that the conformal camera naturally supports this asymmetry.

Recently, building on projective Fourier analysis of the conformal camera, a
mathematical model integrating the head, eyes, and visual cortex into a single
computational binocular system was introduced in \cite{Turski6} with
particular focus on stereopsis. Here it is demonstrated that this integrated
system may efficiently process visual information during fast scanning eye
movements called saccades, employed to build up understanding of a scene
despite the highest acuity only present in the central foveal region of a $2$
deg visual angle. We make about three saccades per second at the eyeball's
maximum speed of $700$ deg/sec. Visual sensitivity is markedly reduced during
saccades as we do not see moving retinal images. These fragmented pieces of
visual information are sent to the cortical areas, with a minor part going to
subcortical areas where they are integrated into a stable coherent percept of
a $3$D world despite of the persistance of incisive eyes movements. This
constancy of vision is maintained by a widespread neural network with multiple
mechanisms receiving inputs from several sources. Not surprisingly, in spite
of a significant recent progress, how this problem is solved by the brain has
been the topic of many theories, see \cite{Wurtz} for a recent review.

The modeling presented in this article, first proposed in \cite{Turski7},
utilizes basic properties of PFT to capture some of the very first
computational aspects of the neural processes during the saccadic eye
movements. First, because the PFT of an image can be efficiently computed by
FFT in complex logarithmic coordinates that also approximate the retinotopy,
the output from the inverse PFT resembles the cortical representation of the
image. Second, a simple translation in retinotopic (logarithmic) coordinates
that is efficiently modeled here by the standard shift property of the inverse
PFT when expressed in these coordinates, remaps the presaccadic scene in the
reference frame centered on the fovea into a postsaccadic reference frame
centered on the impending saccade target. Equivalently, it uniformly shifts
images around the target in cortical periphery to the cortical foveal
location. Moreover, this shift that takes place in retinotopic (logarithmic)
coordinates accounts for perceptual space compression seen around the time of
saccadic eye movements by human subjects in psychophysical laboratory
experiments \cite{Kaiser,Ross}.

The idea of remapping is supported by the fact that the neural correlates of a
copy of\ the oculomotor command to move eyes, known as efference copy or
corollary discharge \cite{Grusser}, have been found in the form of a neuronal
receptive field shift about $50$ ms before a saccade onset in various
retinotopically organized visual cortical areas \cite{Duhamel,Krekelberg}.
This shift points to the possibility that prior to the eyes arriving at the
target, the brain has access to visual information from that peripheral
region. In fact, in the recent experiment \cite{Hunt}, when human subjects
shifted fixation to the clock, their reported time was earlier than the actual
time on the clock by about $40$ ms. It may integrate visual information from
an object across saccades, and therefore, eliminate the need for starting
visual information processing anew three times per second at each fixation and
speed up a costly process of visual information acquisition \cite{VanRullen}.
It may also build up perceptual continuity across fixations \cite{Melcher}.

The conformal camera was initially constructed for the purpose of developing
projectively adapted image representation in the framework of the only well
understood `projective' Fourier analysis formulated as a direction in the
representation theory of semisimple Lie groups, a great achievement of the
$20$th-century mathematics \cite{herb}. In the case of the conformal camera,
it is the representation theory of the group $\mathbf{SL}(2,\mathbb{C})$, the
group generating image projective transformations in a conformal geometry
setting; see \cite{Turski4} where a brief introduction to the group
representations is also given. When writing this article, it became apparent
that we should carefully set a stage for our modeling that involves conformal
projective geometry, abstract and computational harmonic analysis, image
processing, and computational vision including visual neuroscience and machine
vision. Thus, the overarching aim of this article is to elucidate how these
cross-disciplinary areas interwove in our modeling of primate perception.

To this end, the paper is organized as follows. In the next section, we
include, in some detail, physiological and behavioral facts that underlie the
neural processes of human vision related to our computational modeling. In the
following three sections, we lay down the background that explains the
mathematical tools we use in modeling human vision processes. In Section 3, we
introduce the conformal camera and discuss the image projective
transformations. We end this section with the construction of the group of
image projective transformations in the conformal camera. In Section 4, we
review the geometry underlying the conformal camera and demonstrate that the
fundamental properties of this geometry should be uniquely useful in the
early- and intermediate-level vision computational aspects of natural scene
understanding. In the last of these three sections, Section 5, we show that
the conformal camera possesses its own harmonic analysis---projective Fourier
analysis---which gives efficient image representation well adapted to both the
retinal image projective transformations and the retinotopy of the brain's
visual pathways. Finally, in Section 6, we discuss some implementation issues
when working with the discrete PFT. In particular, the binocular system with
head, eyes, and visual cortex numerically integrated by PFT is discussed.
Further, using this integrated binocular system, we model the perisaccadic
perception, including the perisaccadic mislocalizations observed in
psychophysical laboratory experiments. This perisaccadic mislocalization, in
the form of perceptual space compression around the saccade target, is
simulated in the model by the standard shift property of Fourier transform.
Also, the future direction in advancing our modeling and its implementation
are discussed. The paper is summarized in the last section.

The research program presented here advances our mathematical modeling
intended for computational vision, including visual neuroscience and machine
vision systems. It is guided by a strategy important in the contemporary
neurocomputing research: linking known anatomical and physiological details
with efficient computational modeling and engineering designs should be vital
not only to the emerging field of neural engineering but also to interpreting
relevant neurophysiological data.

\section{Visual Neuroscience Background}

\subsection{Visual Perception is a Creative Process}

When light reflected from objects in the $3$D world is impinged upon the
retina, it activates the neuronal pathways, beginning with phototransduction
by about $125$ million photoreceptors. Next, the visual information passes
through a multi-layered circuitry of the retina where substantial processing
takes place.

The only recently emerging picture \cite{Field} of the retinal processing
tells us that more than a dozen of distinct visual recordings of the retinal
image are extracted. For example, one recording emphasizes the boundaries
between objects while another carries information about movement in specific
directions. The result is that more than a dozen of the most essential
features of the original retinal image are extracted in parallel and sent to
the brain as a train of spikes along about $1.5$ million axons of ganglion
cells to more than $30$ association cortex areas containing about $30$ billion
neurons where the details: depth, texture, color, form, motion, etc., are
added and integrated into a coherent view of the $3$D world. This integration
is entirely dependent upon visual experience; almost all higher order features
of vision are influenced by expectations based on past experience. Although
such influences occasionally allow the brain to be fooled into misperception,
as is the case with the optical illusion in Fig.1, they also give us the
ability to see and respond to the visual world quickly.%
\begin{figure}
[h]
\begin{center}
\includegraphics[
natheight=1.881100in,
natwidth=2.992400in,
height=1.8811in,
width=2.9924in
]%
{/document/graphics/1b__1.jpg}%
\caption{This illusion created by Adelson illustrates how perception may
reflect the complex properties of the environment.}%
\end{center}
\end{figure}

We see from this very brief description that visual perception is a creative
process and, for this one reason alone, its quantitative modeling must be
extremely difficult. Therefore, we try to develop a model that captures only
some of the very first computational aspects of visual perception that takes
place the first seconds following the opening of our eyes in daylight. Even
with this limited goal, we find that those aspects are controlled by extremely
sophisticated neural processes that involve nearly every level of the brain.

\subsection{Early Visual Pathways}

When humans open eyes in daylight and direct their gazes to attend a scene,
they only see with the highest clarity, the central part of about a visual
angle of $2$ deg. This region is projected onto the central fovea where its
image is sampled by the hexagonal mosaic of photoreceptors consisting of
mainly cone cells that are color-selective type of photoreceptors for a sharp
daylight vision. The visual acuity decreases rapidly away from the fovea
because the distance between cones increases with eccentricity as they are
outnumbered by rode cells, photoreceptors for a low acuity black-and-white
night vision. Moreover, there is a gradual loss of hexagonal regularity of the
photoreceptor mosaic. For example, at $2.5$ deg radius, \ which corresponds to
the most visually useful region of the retina, acuity drops $50\%$.

The distribution of axons in the optic nerve, which carries the retinal
processing output to the brain, is precisely organized, but varies along the
visual pathways. One aspect of this organization, or the retinotopy\textbf{,
}is that axons corresponding to neighboring places in the retina are
positioned closely in the nerve bundle, with notable exception along the
vertical meridian. This exception stems from the fact that the output of each
eye splits along the retinal vertical meridian when the axons originating from
the nasal half of the retina cross at the optic chiasm to the contralateral
brain's hemisphere and join the temporal half, which remains on the same side
of its eye-of-origin. This splitting and crossing re-organizes the retina
outputs so that the left hemisphere destinations receive information from the
right visual field, and the right hemisphere destinations receive information
from the left visual field. According to the split theory
\cite{Lavidor,Martina}, which provides a greater understanding of vision
cognitive processes than the bilateral theory of overlapping projections,
there is a sharp foveal split along the vertical meridian of hemispherical
cortical projections. Although it is crucial for synthesizing $3$D
representation from the binocular disparities in the pair of $2$D retinal
images, it presents a challenge in modeling retino-cortical image processing
across visual hemifields.

\subsection{Beyond Early Visual Pathways: Visuo-Saccadic Perception}

One of the most important functions of any nervous system is sensing the
external environment and responding in a way that maximizes immediate survival
chances. For this reason, the perception and action have evolved in mammals by
supporting each other's functions. This functional link between visual
perception and oculomotor action is well demonstrated in primates when they
execute the eye-scanning movements (saccades) in order to overcome the eye's
acuity limitation in building up the scene understanding (see Fig. 2).%
\begin{figure}
[h]
\begin{center}
\includegraphics[
natheight=3.241600in,
natwidth=3.854500in,
height=3.2416in,
width=3.8545in
]%
{/document/graphics/2b__2.jpg}%
\caption{(a) San Diego skyline and harbor. (b) Progressively blurred image
from (a) simulating the progressive loss of retinal acuity with eccentricity.
The circle $C_{1}$ encloses the part of the scene projected onto the high
acuity fovea of a $2$ deg diameter. The circle $C_{2}$ encloses the part
projected onto the visually useful faveal region of a $5$ deg diameter. (c) A
scanning path the eyes may take to build the scene understanding. \ Adapted
from \cite{anstis}.}%
\end{center}
\end{figure}

The saccadic eye movement is the most common bodily movement since we make
about three saccades per second at the eyeball's maximum speed of $700$
deg/sec. The eyes remain relatively still (while undergoing tremors, drifts
and microsaccades---a miniature, random eye movement important for proper
functioning of eyes \cite{Martinez-Conde}) between consecutive saccades for
about $180$-$320$ ms, depending on the task performed. During this time
period, the image is processed by the retinal circuitry and sent mainly to the
visual cortex (starting with the primary visual cortex, or V1, and reaching
higher cortical areas, including cognitive areas) with a minor part going to
oculomotor midbrain areas.

The sequence of saccades, fixations, and, often, also smooth-pursuit eye
movements for tracking a slowly moving small object in the scene, is called
the scanpath, first studied in \cite{yarbus}. In Fig. 2, (b) shows a
progressively blurred image from (a), simulating the progressive loss of
acuity with eccentricity. In Fig. 2 (c) we depict the scanpath that eyes might
actually take to build up understanding of the scene.

Although they are the simplest of bodily movements, the eyes' saccades are
controlled by widespread neural network that involves nearly every level of
the brain. Most prominently, it includes the superior colliculus (SC) of the
midbrain for representing possible saccade targets, the parietal eye field
(PEF) and frontal eye field (FEF) in the parietal and frontal lobes of the
neocortex (which obtain inputs from many visual cortical areas) for assisting
the SC in the control of the involuntary (PEF) and voluntary (FEF) saccades.
They also project to the simple neural circuits in the brainstem reticular
formation in the midbrain that ensure the saccade's outstanding speed and precision.

Remarkably, many of the neural processes involved in saccade generation and
control are amenable to precise quantitative studies such that even questions
regarding the operation of the whole structure can be addressed by building on
the existing models \cite{Girard}. This not only carries immense clinical
significance \cite{Carpenter}, but also forms an essential \ preliminary stage
in building our understanding of human vision, the knowledge that will
eventually be transferred to the emerging field of neural engineering.

Nevertheless, some neural processes of the visuo-saccadic system remain
virtually unknown. Visual sensitivity is markedly reduced during saccadic
movements as we do not see moving images on the retinas. This barely
understood neural process is known as saccadic suppression. There is
accumulating evidence that viewers integrate information poorly across
fixations during tasks such as reading, visual search, and scene perception
\cite{Najemnik}. It means that, three times per second, there are instant
large changes in the retinal images without almost any information consciously
carried between images. Furthermore, because the next saccade target selection
for the voluntary saccades takes place in the higher cortical areas involving
cognitive processes \cite{Glimcher}, the time needed for the oculomotor system
to plan and execute the saccadic eye movement could take as long as $150$ ms.
Therefore, it is critical that visual information is efficiently acquired
during each fixation period of about $300$ ms without repeating much of the
whole process at each fixation since it would require too much computational
resources. However, visual constancy, the fact that we are not aware of any
discontinuity in the scene perception when executing the scanpath, is not
perfect. About $50$ ms before the onset of the saccade, during saccadic
movement ($\sim30$ ms) and about $50$ ms after the saccade, perceptual space
is transiently compressed around the saccade target \cite{Kaiser,Ross}, a
phenomenon called perisaccadic mislocalization. We continue this discussion in
Section 6.5 where we present our modeling of the perisacccadic perception
based on projective Fourier transform of the conformal camera.

\section{The Conformal Camera}

We model the human eyes' imaging functions with the conformal camera, the name
of which will be explained later. The camera has many remarkable properties,
the first following directly from its construction: the group of image
projective transformations in the conformal camera is generated internally and
has the `minimal' property as explained in Fig. 3. In the remaining pages of
this article, the other properties will be carefully examined in their
relation to many computational aspects of visual perception.%
\begin{figure}
[h]
\begin{center}
\includegraphics[
natheight=1.658800in,
natwidth=4.356200in,
height=1.6588in,
width=4.3562in
]%
{/document/graphics/3b__3.jpg}%
\caption{(a) Image projective transformations are generated by iterations of
transformations covering translations `$h$' and rotations `$k$' of
\textbf{planar} objects in the scene. (b) The 2D section of the conformal
camera further explains how image projective transformations are generated and
how the projective degrees of freedom are reduced in the camera; one image
projective transformation in the conformal camera corresponds to different
planar objects translations and rotations in the 3D world. }%
\end{center}
\end{figure}

In the conformal camera, the retina is represented by the image plane
$x_{2}=1$ with complex coordinates $x_{3}+ix_{1}$, on which a $3$D scene is
projected under the mapping
\begin{equation}
j(x_{1},x_{2},x_{3})=\left(  x_{3}+ix_{1}\right)  /x_{2}. \label{proj}%
\end{equation}
The implicit assumption $x_{2}\neq0$ will be removed later. Next, we give the
precise form of the `$k$' and `$h$' image transformations introduced in Fig. 3.

\subsection{Basic Image Transformations}

The image projective transformations in the conformal camera are generated by
the following two transformations: (1) an image is projected by $\left(
j|_{\mathbf{S}_{(0,1,0)}^{2}}\right)  ^{-1}$ into the unit sphere
$S_{(0,1,0)}^{2}$ centered at $(0,1,0)$, then the sphere is rotated and the
(rotated) image is projected by $j$ back to the image plane, (2) the image is
translated out of the image plane then projected by $j$ back to the image
plane. The (1) and (2) transformations result in the `$k$'\ and `$h$%
'\ mappings in Fig. 3, respectively. They are explicitly given as follows:

1. $k$\ \textbf{transformations}: $\mathbf{SU(}2\mathbb{)}=\left\{
\binom{\alpha\quad\beta}{-\overline{\beta}\quad\overline{\alpha}}\right\}  $
is the maximal compact subgroup in $\mathbf{SL(}2,\mathbb{C)}$, the group of
$2\times2$ complex matrices of determinant $1$. We let the group
$\mathbf{SO(}3\mathbb{)}$ of three dimensional rotations act on the sphere
$S_{(0,1,0)}^{2}$ by rotating it about $(0,1,0)$. Furthermore, we parametrize
$\mathbf{SO(}3\mathbb{)}$ by the Euler angles $(\psi,\phi,\psi^{\prime})$,
where $\psi$ is the rotation about the $x_{2}$-axis, followed by the rotation
$\phi$ about the $y_{3}$-axis, whichis parallel to the $x_{3}$-axis and passes
through $(0,1,0)$, and finally by the rotation $\psi^{\prime}$ about the
rotated $x_{2}$-axis. Then, to each $R(\psi,\phi,\psi^{\prime})$ in
$\mathbf{SO(}3\mathbb{)}$ there correspond two elements in $\mathbf{SU(}%
2\mathbb{)}$,%
\begin{equation}
k(\psi,\phi,\psi^{\prime})=\pm\left(
\begin{array}
[c]{cc}%
e^{i(\psi+\psi^{\prime})/2}\cos\frac{\phi}{2} & ie^{i(\psi-\psi^{\prime}%
)/2}\sin\frac{\phi}{2}\\
ie^{-i(\psi-\psi^{\prime})/2}\sin\frac{\phi}{2} & e^{-i(\psi+\psi^{\prime}%
)/2}\cos\frac{\phi}{2}%
\end{array}
\right)  \text{,} \label{k}%
\end{equation}
such that\textbf{\ } $j\circ R(\psi,\phi,\psi^{\prime})\circ\left(
j|_{\mathbf{S}_{(0,1,0)}^{2}}\right)  ^{-1}(z)=k\cdot z$ \ are given by the
following linear fractional mappings%

\begin{equation}
k(\psi,\phi,\psi^{\prime})\cdot z=\frac{(e^{-i(\psi+\psi^{\prime})/2}\cos
\frac{\phi}{2})\text{ }z+ie^{i(\psi-\psi^{\prime})/2}\sin\frac{\phi}{2}%
}{(ie^{i(\psi-\psi^{\prime})/2}\sin\frac{\phi}{2})\text{ }z+e^{i(\psi
+\psi^{\prime})/2}\cos\frac{\phi}{2}}\text{.} \label{lfk}%
\end{equation}

2. $h$ \textbf{transformations: }Similarly, for each translation vector
$\overrightarrow{b}=(b_{1},b_{2},b_{3})$ where $b_{2}\neq-1$ acting on the
image plane $T_{\overrightarrow{b}}(x)=x+$ $\overrightarrow{b}$, there are two
elements $\mathbf{SL}(2,\mathbb{C})$,
\begin{equation}
h(b_{1,}b_{2},b_{3})=\pm\left(
\begin{array}
[c]{cc}%
\left(  1+b_{2}\right)  ^{1/2} & 0\\
(b_{3}+ib_{1})\left(  1+b_{2}\right)  ^{-1/2} & \left(  1+b_{2}\right)
^{-1/2}%
\end{array}
\right)  \label{h}%
\end{equation}
such that $j\circ T_{\overrightarrow{b}}\circ\left(  j|_{x_{2}=1}\right)
^{-1}(z)=h\cdot z$ are given by the corresponding linear fractional mappings
by the same action as before,
\begin{equation}
h(b_{1},b_{2},b_{3})\cdot z=\frac{\left(  1+b_{2}\right)  ^{-1/2}%
z+(b_{3}+ib_{1})\left(  1+b_{2}\right)  ^{-1/2}}{\left(  1+b_{2}\right)
^{1/2}}. \label{lfh}%
\end{equation}

Now, if $f(z)$ is an image intensity function and $g$ is either $k$ or $h$
mapping, the corresponding image transformation is the following:
$f(g^{-1}\cdot z)$.

Both $k\cdot z$ and $h\cdot z$ mappings have forms of special
linear-fractional transformations%
\[
g\cdot z=\frac{\alpha z+\beta}{\gamma z+\delta};\text{\quad}\alpha
\delta-\gamma\beta=1.
\]

These mappings are conformal, that is, they preserve the oriented angles of
two tangent vectors $z_{k}^{\prime}(t_{0})$ to any two curves $z_{k}(t)$
($k=1,2$) intersecting at the point $q=z(t_{0})$. In fact,
\begin{equation}
\frac{d}{dt}\left(  \frac{\alpha z_{k}(t)+\beta}{\gamma z_{k}(t)+\delta
}\right)  _{t=t_{0}}=\frac{z_{k}^{\prime}(t_{0})}{(\gamma q+\delta)^{2}}%
=\frac{e^{i\chi(q)}z_{k}^{\prime}(t_{0})}{|(\gamma q+\delta)|^{2}};\quad
k=1,2, \label{conformal k and h}%
\end{equation}
and both vectors $z_{1}^{\prime}(t_{0})$ and $z_{2}^{\prime}(t_{0})$ are
rotated by the same angle $\chi(q)$.

\subsection{The Group of Image Projective Transformations}

\subsubsection{The $\mathbf{PSL}(2,\mathbb{C})$ Group}

The group of image transformations in the conformal camera is generated by all
finite iterations of $k$ and $h$ mappings. To derive this group, we recall
that $k\in\mathbf{SU(}2\mathbb{)}$ and note that $h\in\mathbf{A}%
\overline{\mathbf{N}}\subset$ $\mathbf{SL}(2,\mathbb{C})$ if $1+b_{2}>0$ and
$h=\varepsilon\mathbf{A}\overline{\mathbf{N}}\subset\mathbf{SL}(2,\mathbb{C})$
if $1+b_{2}<0,$ where
\begin{equation}
\mathbf{A}=\left\{  \binom{\rho\quad0}{0\quad\rho^{-1}}\right\}
\text{,\quad\ }\overline{\mathbf{N}}=\left\{  \binom{1\quad0}{\xi\quad
1}\right\}  \text{,\quad\ }\epsilon=\binom{-i\quad0}{0\quad i}.
\label{matrices}%
\end{equation}
Now, it follows from the polar decomposition\ $\mathbf{SL}(2,\mathbb{C}%
)=\mathbf{SU(}2\mathbb{)}\mathbf{ASU(}2\mathbb{)}$, that all these finite
iterations result in the group $\mathbf{SL}(2,\mathbb{C})$ acting by
linear-fractional mappings
\begin{equation}
\mathbf{SL}(2,\mathbb{C})\ni\binom{a\ ~b}{c\text{ \ }d}\cdot z=\frac
{dz+c}{bz+a};\quad z=x_{3}+ix_{1}\equiv(x_{1},1,x_{3}). \label{lft1}%
\end{equation}

Because $\pm\binom{a\quad b}{c\quad d}$ have the same action, we need to
identify matrices in $\mathbf{SL}(2,\mathbb{C})$ that differ in sign. The
result is the quotient group $\mathbf{PSL(}2,\mathbb{C)}=\mathbf{SL(}%
2,\mathbb{C)}/\{\pm Id\}$, where $Id$ is the identity matrix, and the action
(\ref{lft1}) establishes a group isomorphism between linear-fractional
mappings and $\mathbf{PSL(}2,\mathbb{C)}$. Thus,%
\begin{equation}
\mathbf{PSL}(2,\mathbb{C})\ni g=\binom{a\ ~b}{c\text{ \ }d}\longmapsto
f\left(  g^{-1}\cdot z\right)  =f\left(  \frac{az-c}{-bz+d}\right)
\label{IPT}%
\end{equation}
gives the image projective transformations of the intensity function $f(z)$.

\subsubsection{Conformality}

As we showed in (\ref{conformal k and h}), the mappings in (\ref{lft1}) are
conformal. Because of this property, the camera is called `conformal'.
Although, the conformal part of an image projective transformation can be
removed with almost no computational cost, leaving only a perspective
transformation of the image (see \cite{Turski4,Turski5}); the conformality
provides an advantage in imaging because the conformal mappings rotate and
dilate the image infinitesimal neighborhoods, and, therefore, locally preserve
the image `pixels'.

To complete the description of the conformal camera, we need to address some
implicit assumptions, such as the restriction $-bz+d\neq0$ in (\ref{IPT}) we
have frequently made in this section.

\section{Geometry of the Conformal Camera}

In the homogeneous coordinate framework of projective geometry \cite{berger},
the conformal camera is embedded into the complex plane%
\[
\mathbb{C}^{2}=\left\{  \binom{z_{1}}{z_{2}}\mid z_{1}=x_{2}+iy,z_{2}%
=x_{3}+ix_{1}\right\}  .
\]
In this embedding, the `slopes' $\xi$ of the complex lines $z_{2}=\xi z_{1}$
are numerically identified with the points on the extended image
plane\textbf{\ }$\widehat{\mathbb{C}}=\mathbb{C\cup\{\infty)}$ where $\infty$
corresponds to the line $z_{1}=0$. We note that if $x_{2}\neq0$ and $y=0$, the
slope\ $\xi$ corresponds to the point $x_{3}+ix_{1}$ at which the ray (line)
in $\mathbb{R}^{3}$ that passes through the origin is intersecting the image
plane of the conformal camera.

Now, the standard action of the group $\mathbf{SL(}2,\mathbb{C)}$ on nonzero
column vectors $\mathbb{C}^{2}$,%
\[
\left(
\begin{array}
[c]{c}%
z_{1}^{\prime}\\
z_{2}^{\prime}%
\end{array}
\right)  =\left(
\begin{array}
[c]{cc}%
a & b\\
c & d
\end{array}
\right)  \left(
\begin{array}
[c]{c}%
z_{1}\\
z_{2}%
\end{array}
\right)  =\left(
\begin{array}
[c]{c}%
az_{1}+bz_{2}\\
cz_{1}+dz_{2}%
\end{array}
\right)
\]
implies that the slope $\xi=\frac{z_{2}}{z_{1}}$ is mapped to the slope
\[
\xi^{\prime}=\frac{z_{2}^{\prime}}{z_{1}^{\prime}}=\frac{cz_{1}+dz_{2}}%
{az_{1}+bz_{2}}=\frac{c+d\xi}{a+b\xi}%
\]
agreeing with the linear fractional mappings in (\ref{lft1}).

However, the action must be extended to include the line $z_{1}=0$ of `slope'
$\infty$ as follows:%
\[
\binom{a~b}{c\text{ }d}\cdot\infty=d/b,\ \ \binom{a~b}{c\text{ }d}%
\cdot(-a/b)=\infty\text{.}%
\]
The stereographic projection $\sigma=j|_{S_{(0,1,0)}^{2}}$ (with $j$ in
(\ref{proj})) maps $S_{(0,1,0)}^{2}$ bijectively onto \textbf{\ }%
$\widehat{\mathbb{C}}$ and $\sigma(0,0,0)=\infty$ gives a concrete meaning to
the point $\infty$ such that it can be treated as any other point of
$\widehat{\mathbb{C}}$. Thus, geometry of the image plane $\widehat
{\mathbb{C}}$ of the conformal camera with the image projective
transformations given by the group $\mathbf{PSL(}2,\mathbb{C)}$ acting by
linear-fractional transformations can be dually described as follows:

1. $\widehat{\mathbb{C}}$ is the complex projective line, i.e., $\widehat
{\mathbb{C}}\cong P^{1}\left(  \mathbb{C}\right)  $ where
\[
P^{1}\left(  \mathbb{C}\right)  =\left\{  \text{complex lines in }%
\mathbb{C}^{2}\text{ through the origin}\right\}
\]
with the group of projective transformations $\mathbf{PSL(}2,\mathbb{C)}$.
Thus, the image projective transformations acting on the points of the
extended image plane (or simply, the image plane) of the conformal camera can
be identified with projective geometry (containing Euclidean geometry as \ a
sub-geometry) of the one-dimensional complex line \cite{berger}.

2. $\widehat{\mathbb{C}}$ is the Riemann sphere since under stereographic
projection $\sigma=j|_{S_{(0,1,0)}^{2}}$ we have the isomorphism
$\widehat{\mathbb{C}}\cong S_{(0,1,0)}^{2}.$ The group $\mathbf{PSL(}%
2,\mathbb{C)}$ acting on $\widehat{\mathbb{C}}$ consists of the bijective
meromorphic mappings of $\widehat{\mathbb{C}}$ \cite{jones}. Thus, it is the
group of holomorphic automorphisms of the Riemann sphere that preserve the
intrinsic geometry imposed by complex structure, known as M\"{o}bius geometry
\cite{henle} or inversive geometry\textit{\ }\cite{brannan}.

What we have just described shows the following fundamental property:
projective geometry underlying the conformal camera, also called M\"{o}bius or
inversive geometry, and holomorphic complex structure that provides the
framework for the development of complex numerical analysis, are in fact two
faces---one `geometric' and the other `numerical'---of the same coin. We
stress that the real projective geometry underlying the pinhole camera and
usually employed in computer vision \cite{mundy,shapiro} does not possess this
fundamental property which sets apart our modeling of primate visual
perception from other approaches.

\subsection{The Conformal Camera and Visual Perception}

The image plane of the conformal camera does not admit a distance that is
invariant under image projective (that is, linear-fractional) transformations.
Therefore, geometry of the conformal camera does not possess a Riemannian
metric; for instance, there is no curvature measure. As customary in complex
projective (M\"{o}bius or inversive) geometry, we consider a line as a circle
passing through the point $\infty.$ Then, the fundamental property of this
geometry can be expressed as follows: linear-fractional mappings take circles
to circles. Thus, circles can play the role of geodesics. Moreover, each
circle carries a signature of curvature---the inverse of the radius. We showed
before that linear-fractional mappings are conformal; we add here for
completeness that stereographic projection $\sigma=j|_{S_{(0,1,0)}^{2}}$ is
also conformal and maps circles in the sphere $S_{(0,1,0)}^{2}$ onto circles
in $\widehat{\mathbb{C}}$. In conclusion, circles play a crucial role in the
conformal camera geometry and it should be reflected in psychological and
computational aspects of natural scene understanding if this camera is
relevant to modeling primate visual perception.

Neurophysiological experiments demonstrate that the retina performs filtering
of impinged images that extract local contrast spatially and temporally. For
instance, center surround cells at the retinal processing stage are triggered
by local spatial changes in intensity referred to as edges or contours. This
filtering is enhanced in the primary visual cortex, the first cortical area
receiving, via LGN, the retinal output, which itself is a case study in dense
packing of overlapping visual submodalities: motion, orientation, frequency
(color), and oculomotor dominance (depth). In psychological tests, humans
easily detect a significant change in spatial intensity (low-level vision),
and effortlessly and unambiguously group this usually fragmented visual
information (contours of occluded objects, for example), into coherent, global
shapes (intermediate-level vision). Considering its computational complexity,
it is one of the most difficult problems that primate visual system has to
solve \cite{ullman}.

The Gestalt phenomenology and quantitative psychological measurements
established the rules, summarized in the ideas of good continuation
\cite{kofka,wertheimer} and association field \cite{field}, that determine
interactions between fragmented edges such that they extend along continuous
contours joining them in the way they will normally be grouped together to
faithfully represent a scene. Evidence accumulated in psychological and
physiological studies suggests that the human visual system utilizes a local
grouping process (association field) with two very simple rules: collinearity
and co-circularity with underlying scale invariant statistics for both
geometric arrangements in natural scenes. These rules were confirmed in
\cite{sigman,chow} by statistical analysis of natural scenes. Two basic
intermediate-level descriptors that the brain employs in grouping elements
into global objects are the medial axis transformation \cite{blum}, or
symmetry structure \cite{leyton1,leyton2}, and the curvature extrema
\cite{attneave,hoffman}. In fact, the medial axis, which visual system
extracts as a skeletal (intermediate-level) representation of objects
\cite{kovacs}, can be defined as the set of the centers of maximal circles
inscribed inside the contour. The curvatures at the corresponding points of a
contour are given by the inverse radii of the circles.

From the above discussion we see that, on one hand, co-circularity and scale
invariance emerge as the most basic concepts used by intermediate-level vision
in solving the difficult problems of grouping local elements into individual
objects of natural scenes. On the other hand, the non-metric projective
geometry of the conformal camera that models eye imaging functions can be
entirely constructed from circles such that co-circularity is preserved by
projective transformations. Thus, it seems that the conformal camera would be
very useful in modeling eye's imaging functions related to the lower and
intermediate-level natural vision.

Other characteristics of the conformal camera that are\ uniquely useful in
modeling primate visual perception are discussed in the remaining part of this
article. Next, we briefly review the unity of geometry and numerical methods
by showing that the conformal camera has its own projective Fourier transform (PFT).

\section{Projective Fourier Analysis}

The projective Fourier analysis has been constructed by restricting geometric
Fourier analysis of $\mathbf{SL(}2,\mathbb{C)}$---a direction in the
representation theory of the semisimple Lie groups \cite{knapp}---to the image
plane of the conformal camera (see Section 5.1 in \cite{Turski5}). The
resulting projective Fourier transform (PFT)\textit{\ }of a given image
intensity function $f(z)\in L^{2}(\mathbb{C})$ is the following
\begin{equation}
\widehat{f}(s,k)=\frac{i}{2}\int f(z)|z|^{-is-1}\left(  \frac{z}{|z|}\right)
^{-k}dzd\overline{z} \label{PFT}%
\end{equation}
where $(s,k)\in\mathbb{R\times Z}$ and if $z=x_{3}+ix_{1}$, then $\frac{i}%
{2}dzd\overline{z}=dx_{3}dx_{1}$ is the Euclidean measure on the image plane.
In this work, we consider only the noncompact picture of PFT and, for a
complete mathematical account, that includes also the compact picture we refer
to \cite{Turski5}. The noncompact and compact pictures in the case of
Euclidean group correspond to the classical and spherical Fourier analyses,
respectively (Section 3 in \cite{Turski4}). In the next remark we justify the
name `projective Fourier transform', and, for comprehensive account, we refer
to \cite{Turski5}.

\begin{remark}
The functions $\Pi_{s,k}(z)=|z|^{is}\left(  \frac{z}{|z|}\right)  ^{k}$;
$s\in\mathbb{R}$, $k\in\mathbb{Z}$ are all one dimensional unitary
representations of the Borel subgroup $\mathbf{B=MAN}$ of $\mathbf{SL(}%
2,\mathbb{C)}$, and they play in (\ref{PFT}) the role complex exponentials
play in the classical Fourier transform. \ These one dimensional
representations are all finite unitary representations of the Borel subgroup
$\mathbf{B}$, as opposed to the fact that all nontrivial unitary
representations of $\mathbf{SL(}2,\mathbb{C)}$ are infinite. Furthermore, the
group $\mathbf{B}$ `exhausts'\ the projective group $\mathbf{SL(}%
2,\mathbb{C)}$ by Gauss decomposition $\mathbf{SL(}2,\mathbb{C)}\overset
{.}{\mathbb{=}}\overline{\mathbf{N}}\mathbf{B}$, where `$\overset
{.}{\mathbb{=}}$'\ means that the equality holds up to lower dimensional
subset, that is, almost everywhere, and $\overline{\mathbf{N}}$ in
(\ref{matrices}) represents Euclidean translations.
\end{remark}

In log-polar coordinates $(u,\theta)$ given by $\ln re^{i\theta}=\ln
r+i\theta=u+i\theta$, $\widehat{f}(k,s)$ has the form of the standard Fourier
integral
\begin{equation}
\widehat{f}(s,k)=\int\int f(e^{u+i\theta})e^{u}e^{-i(us+\theta k)}%
dud\theta\text{,} \label{sft}%
\end{equation}
where we used $\frac{i}{2}dzd\overline{z}=e^{2u}dud\theta$. We see that a
function $f$ that is integrable on $\mathbb{C}^{\ast}=\mathbb{C}%
\backslash\{0\}$, has finite PFT,
\begin{equation}
\left\vert \widehat{f}(s,k)\right\vert \leq\int_{0}^{2\pi}\int_{-\infty
}^{u_{1}}f(e^{u+i\theta})e^{u}dud\theta=\int_{0}^{2\pi}\int_{0}^{r_{1}%
}f(re^{i\theta})drd\theta<\infty. \label{Ineq}%
\end{equation}
Therefore, this $f$ can be extended to $\mathbb{C}$ by $f(0)=0.$ Thus, in
spite of the logarithmic singularity of log-polar coordinates, the projective
Fourier transform of integrable functions on $\mathbb{C}$ is finite. This
observation will be crucial when we discretize the PFT in the next section.

Inverting (\ref{sft}), which is done in the $(u,\theta)$-space, we get
\begin{equation}
e^{u}\mathrm{f}(u,\theta)=\frac{1}{\left(  2\pi\right)  ^{2}}\sum_{k=-\infty
}^{\infty}\int\widehat{f}(s,k)e^{i(us+\theta k)}ds\text{,} \label{isft}%
\end{equation}
where $\mathrm{f}(u,\theta)=f(e^{u+i\theta})$. We stress that although
$f(e^{u+i\theta})$ and $\mathrm{f}(u,\theta)$ are numerically equal, they are
given on different spaces; $f(e^{u+i\theta})$ is on the image plane in polar
coordinates and $\mathrm{f}(u,\theta)$ is on the space defined by rectangular
$(u,\theta)$-coordinates.

Finally, by expressing (\ref{isft}) in the $z$-variable, we obtain the inverse
projective Fourier transform%
\begin{equation}
f(z)=\frac{1}{\left(  2\pi\right)  ^{2}}\sum_{k=-\infty}^{\infty}\int
\widehat{f}(s,k)|z|^{is-1}\left(  \frac{z}{|z|}\right)  ^{k}ds. \label{IPFT}%
\end{equation}

\subsection{Discrete Projective Fourier Transform}

To discretize the PFT we use the fact that $\widehat{f}(s,k)$ is finite for an
integrable function $f$, see (\ref{Ineq}). By removing a disk $|z|\ \leq
r_{a}$, we can assume that the support of $\mathrm{f}(u,\theta)$ is contained
within $(\ln r_{a},\ln r_{b})\times\lbrack0,2\pi)$. We approximate the
integral in (\ref{sft}) by a double Riemann sum
\[
\widehat{f}\left(  2\pi m/T,n\right)  \approx\frac{2\pi T}{NM}\sum_{k=0}%
^{M-1}\sum_{l=0}^{N-1}e^{u_{k}}f(e^{u_{k}}e^{i\theta_{l}})e^{-2\pi
i(mk/M+nl/N)}%
\]
with $M\times N$ partition points
\begin{equation}
(u_{k},\theta_{l})=\left(  \ln r_{a}+kT/M,2\pi l/N\right)  ;0\leq k\leq
M-1,0\leq l\leq N-1,\text{ }T=\ln(r_{b}/r_{a})\text{.} \label{l-p discret.}%
\end{equation}
Then, introducing%
\begin{equation}
f_{k,l}=(2\pi T/MN)f(e^{u_{k}}e^{i\theta_{l}})\text{ and }\mathrm{f}%
_{k,l}=(2\pi T/MN)\mathrm{f}(u_{k},\theta_{l}) \label{pixels}%
\end{equation}
and defining $\widehat{f}_{m,n}$ by $^{{}}$%
\begin{equation}
\widehat{f}_{m,n}=\sum_{k=0}^{M-1}\sum_{l=0}^{N-1}f_{k,l}e^{u_{k}}e^{-i2\pi
mk/M}e^{-i2\pi nl/N}, \label{lpdft}%
\end{equation}
we obtain%
\begin{equation}
\mathrm{f}_{k,l}=\frac{1}{MN}\sum_{m=0}^{M-1}\sum_{n=0}^{N-1}\widehat{f}%
_{m,n}e^{-u_{k}}e^{i2\pi mk/M}e^{i2\pi nl/N}\text{.} \label{ilpdft}%
\end{equation}

We note that $\widehat{f}_{m,n}\approx\widehat{f}\left(  2\pi m/T,n\right)  $
and refer to \cite{henrichi} for a discussion of numerical aspects on the
approximation. Both expressions (\ref{lpdft}) and (\ref{ilpdft}) can be
computed efficiently by FFT algorithms since the exponents are taken at
equidistant points. See simulation for a bar pattern in Fig. 4.%
\begin{figure}
[h]
\begin{center}
\includegraphics[
natheight=1.617100in,
natwidth=4.182100in,
height=1.6171in,
width=4.1821in
]%
{/document/graphics/4b__4.jpg}%
\caption{Exp-polar sampling (the distance between circles partially displayed
in the first quadrant changes exponentially) of a bar pattern on the retina is
shown on the left. The bar pattern in the cortical space rendered by the
inverse DPFT computed with FFT is shown on the right. The cortical uniform
sampling grid, which is obtained by applying complex logarithm to the
exp-polar grid in (a), is shown only in the upper left corner.}%
\end{center}
\end{figure}

Finally, on introducing $z_{k,l}=e^{u_{k}+i\theta_{l}}$ into (\ref{lpdft}) and
(\ref{ilpdft}), we arrive at the $(M,N)$-point discrete projective Fourier
transform (DPFT)\ and its inverse:
\begin{equation}
\widehat{f}_{m,n}=\sum_{k=0}^{M-1}\sum_{l=0}^{N-1}f_{k,l}\left(  \frac
{z_{k,l}}{|z_{k,l}|}\right)  ^{-n}|z_{k,l}|^{-i2\pi m/T+1} \label{dpft}%
\end{equation}
and
\begin{equation}
f_{k,l}=\frac{1}{MN}\sum_{m=0}^{M-1}\sum_{n-0}^{N-1}\widehat{f}_{m,n}\left(
\frac{z_{k,l}}{|z_{k,l}|}\right)  ^{n}|z_{k,l}|^{i2\pi m/T-1}, \label{idpft}%
\end{equation}
now with $f_{k,l}=(2\pi T/MN)f(z_{k,l})$. The projectively adapted
characteristics\textit{\ }of the discrete projective Fourier analysis can be
expressed as follows:
\begin{equation}
f_{k,l}^{\prime}=\frac{1}{MN}\sum_{m=0}^{M-1}\sum_{n=0}^{N-1}\widehat{f}%
_{m,n}\left(  \frac{z_{k,l}^{\prime}}{|z_{k,l}^{\prime}|}\right)  ^{n}%
|z_{k,l}^{\prime}|^{i2\pi m/T-1}\text{,} \label{dppcc'}%
\end{equation}
where $z_{k,l}^{\prime}=g^{-1}\cdot z_{k,l}$, $g\in\mathbf{SL(}2,\mathbb{C)}$
and $f_{k,l}^{\prime}=(2\pi T/MN)f(z_{k,l}^{\prime})$.

Although projective characteristics must be derived in $z$-coordinates, in
practical image processing, (\ref{dppcc'}) should be expressed in log-polar
coordinates to be fast computable by FFT. To this end, let $(u_{m,n}^{\prime
},\theta_{m,n}^{\prime})$ denote log-polar coordinates of $z_{m,n}^{\prime
}=e^{u_{m,n}^{\prime}}e^{i\theta_{m,n}^{\prime}}$. In these coordinates,
(\ref{dppcc'}) is given by the following expression (see
\cite{Turski4,Turski5} for details)%

\[
\mathrm{f}_{m,n}^{^{\prime}}=\frac{1}{MN}\sum_{{\small k=0}}^{{\small M-1}%
}\sum_{{\small l=0}}^{{\small N-1}}\widehat{f}_{k,l}e^{-u_{m,n}^{\prime}%
}e^{i2\pi u_{m,n}^{\prime}k/T}e^{i\theta_{m,n}^{\prime}lL}.
\]
Thus, we can render image projective transformations in terms of projective
Fourier transform of the original image only.

\section{DPFT in Computational Vision}

We discussed before the relevance of the conformal camera to the
intermediate-level vision task of grouping image elements into individual
objects in natural scenes. Here we want to discuss the relevance of the data
model of image representation based on projective Fourier analysis to image
processing in computational vision, including visual neuroscience and
biologically motivated machine vision systems.

\subsection{Modeling the Retinotopy}

The mappings $w=\ln(z\pm a)-\ln a$, with $a>0$ and $\pm a$ indicating, for
different signs, the left or right brain hemisphere, are accepted
approximations of the topographic structure of primate primary visual cortex
(V1) \cite{Schwartz}, where the parameter $a$ removes the singularity of the
logarithm. However, the discrete projective Fourier transform (DPFT) that
provides the data model for retinal image representation, can be efficiently
computed by FFT only in log-polar coordinates given by the complex logarithm
$w=\ln z$, the mapping with distinctive rotational and zoom symmetries:%
\[
\ln(e^{i\theta}z)=\ln z+i\theta\text{,\quad}\ln(\rho z)=\ln z+\ln\rho\text{.}%
\]
Thus, we see that the Schwartz model of the retina comes with drastic
consequences; it destroys rotation and zoom symmetries. We also recall that
PFT in log-polar coordinates does not have a singularity at the origin, see
(\ref{Ineq}).

The following facts support our modeling with DPFT. First, for small $|z|\ \ll
a$, $\ln(z\pm a)-\ln a$ is approximately linear while, for large $|z|\ \gg a$,
it is dominated by $\ln z$. Secondly, to construct discrete sampling for DPFT,
the image was regularized by removing a disc representing the fovea (see
previous section). Thirdly, there is accumulated evidence pointing to the fact
that the fovea and periphery have different functional roles in vision
\cite{petrov,prado,Xing} and likely involve different image processes.
Finally, by the split theory of hemispherical image representation, which we
mentioned before, the foveal region has a discontinuity along the vertical
meridian, with each half processed in a different brain hemisphere
\cite{Lavidor}. We note that the two hemispheres are connected by a massive
bridge of $500$ million neuronal axons called the corpus callosum.

We conclude this discussion with the following remarks: both models our and
Schwartz' model in \cite{Schwartz} (see Fig. 5), as well as all other similar
models, are, in fact, fovea-less models \cite{weiman}.Furthermore, since the
fovea is explicitly removed in our modeling, we expect to extend the present
model to include foveal representation in the next stage of this modeling. In
fact, the lack of the fovea in our modeling is one of the challenges that is
stalling implementation of the model.%
\begin{figure}
[h]
\begin{center}
\includegraphics[
natheight=1.467600in,
natwidth=4.006500in,
height=1.4676in,
width=4.0065in
]%
{/document/graphics/5b__5.jpg}%
\caption{(a) Schwartz model of the retina: the strip of width $2$a is removed
and two half-maps of $\ln z$ are shifted to meet along the vertical meridian.
(b) Our model: the fovea is removed and the retina is split along the vertical
meridian, conforming to the split theory of the retino-cortical projection.}%
\end{center}
\end{figure}
We continue this discussion in Section 6.5.1.

\subsection{On Numerical Implementation of DPFT}

The DPFT approximation was obtained using the rectangular sampling grid
$(u_{k},\theta_{l})$ in (\ref{l-p discret.}), corresponding, under the
mapping,
\[
w_{k,l}=u_{k}+i\theta_{l}\longmapsto z_{k,l}=e^{u_{k}+i\theta_{l}}%
=r_{k}e^{i\theta_{l}}%
\]
to nonuniform sampling grid with equal sectors
\begin{equation}
\alpha=\theta_{l+1}-\theta_{l}=\frac{2\pi}{N}\text{, }l=0,1,...,N-1
\label{alpha}%
\end{equation}
and with ring radii increasing exponentially
\begin{equation}
\rho_{k}=r_{k+1}-r_{k}=e^{u_{k+1}}-e^{u_{k}}=e^{u_{k}}(e^{\delta}%
-1)=r_{k}(e^{\delta}-1)\text{, }k=0,1,...,M-1, \label{ro}%
\end{equation}
where $\delta=u_{k+1}-u_{k}$. The radii $r_{k}=r_{0}e^{k\delta}$ are given in
terms of the spacing $\delta=\frac{T}{M}$ and $r_{0}=r_{a}$, where $r_{a} $ is
the radius of the disc that has been removed to regularize logarithmic
singularity, see (\ref{l-p discret.}).

Lets assume that we have been given a picture of the size $A\times B$ in pixel
units, which is displayed with $K$ dots per unit length (dpl). Then, the
physical dimensions, in the chosen unit of length, of the pixel and the
picture are $1/K\times1/K$ and $A/K\times B/K$, respectively. Also, we assume
that the retinal coordinates' origin (fixation) is the picture's center.

The central disc of radius $r_{0}$ represents the fovea with a uniformly
distributed of grid points and the number of the foveal pixels $N_{f}$ given
by $\pi r_{0}^{2}=N_{f}/K^{2}$. This means that the fovea cannot increase the
resolution, which is related to the distance of the picture from the eye. The
number of sectors is obtained from the condition $2\pi(r_{0}+r_{1})/2\approx
N(1/K)$, where $N=[2\pi r_{0}K+\pi]$. Here $[a]$ is the closest integer to
$a$. To get the number of rings $M$, we assume that $\rho_{0}=r_{0}(e^{\delta
}-1)=1/K$ and $r_{b}=r_{M}=r_{0}e^{M\delta}$. We can take either
$r_{b}=(1/K)\min(A,B)/2$ or $r_{b}=(1/K)\sqrt{A^{2}+B^{2}}/2 $. Thus,
$\delta=\ln[(1+1/r_{0}K]$ and $M=(1/\delta)\ln(r_{b}/r_{0})$.

\begin{example}
We let $A\times B=512\times512$ and $K=4$ per mm, so the physical dimensions
in mm are $128$ $\times128$ and $r_{b}=128\sqrt{2}/2=90.5$. Furthermore, we
let $N_{f}=296$, so $r_{0}=2.427$ and $N=64$. Finally, $\delta=\ln
(10.7084/9.7084)\approx0.09804$ and $\ (1/0.09804)\ln(90.5/2.427)\approx
M=37.$ The sampling grid consists of points in polar coordinates: $(r_{k}%
+\rho_{k+1}/2,\theta_{l}+\pi/64)=(2.552e^{k0.09804},(2l+1)\pi/64)$
$k=0,1,...,36,l=0,1,...,63$.
\end{example}

In this example, the number of pixels in the original image is $262,144$,
whereas the foveal (uniform sampling) and peripheral (log-polar sampling)
representation of the image contain only $2,664$ pixels.

We stress again that $\mathrm{f}_{k,l}$ and $f_{k,l}$ are discretizations of
the same image in different planes; $f_{k,l}$ are the image samples in the
image plane sampled on a nonuniform grid $\left(  e^{u_{k}}e^{i\theta_{l}%
}\right)  $, while the inverse DPFT output (\ref{ilpdft}) gives the image
samples $\mathrm{f}_{k,l}$ on the uniform grid $\left(  u_{k},\theta
_{l}\right)  $, where $u_{k}=\ln r_{k}$.

In summary, a simple description of the imaging model based on DPFT is as
follows: an image (analog or digital) of a scene impinged on the retina is
sampled on a nonuniform exp-polar grid, $\{r_{k}e^{i\theta_{l}}\}_{M\times N}%
$, that approximates the density distribution of retinal ganglion cells,
giving the set of pixels $\left\{  f_{k,l}\right\}  _{N\times M}$. In this
grid, the radial spacing changes exponentially: $r_{k}=r_{a}e^{\delta k}$,
$k=1,2,...,M$, and the angular spacing is constant: $\theta_{l}=\alpha l$,
$l=1,2,...,N$. As it was shown in Example 2, this sampling results in about
$100$ times less pixels than in the original image. To render $\left\{
f_{k,l}\right\}  _{N\times M}$, the DPFT is formed and computed by FFT in
log-polar coordinates $(u_{k},\theta_{l})$ obtained by applying a complex
logarithm as follows: $\ln(r_{a}e^{\delta k}e^{i\alpha l})=\ln r_{a}+\delta
k+i\alpha l=u_{k}+i\theta_{l}$, resulting in the set $\left\{  \widehat
{f}_{k,l}\right\}  _{M\times N}$. Next, the IPFT is assembled and computed
again by FFT, this time giving the image samples $\mathrm{f}_{k,l}=f_{k,l}$
rendered in cortical (log-polar) coordinates $(u_{k},\theta_{l})$.

\subsection{Relation to Other Numerical Approaches}

From the numerical approaches to foveate (or space-variant) vision, involving,
for example, Fourier-Mellin transform or log-polar Hough transform, the most
closely related to our work are results reported by Schwartz' group at Boston
University. We note that the approximation of the retinotopy by a complex
logarithm (see Section 6.1) was first proposed by Eric Schwartz in 1977. This
group introduced the fast exponential chirp transform (FECT) \cite{bonmassar}
in their attempt to develop numerical algorithms for space-variant image
processing. Basically, both FECT and its inverse were obtained by the change
of variables in both the spatial and frequency domains in the standard Fourier
integrals. The discrete FECT was introduced somehow ad hoc, without references
to numerical aspects of the approximation. Moreover, some basic components of
Fourier analysis, such as underlying geometry or Plancherel measure was not
considered. In comparison, projective Fourier transform (PFT) provides an
efficient image representation well adapted to projective transformations
produced in the conformal camera by the group $\mathbf{SL(}2,\mathbb{C)}$
acting on the image plane by linear-fractional mappings. Significantly, PFT
can be obtained by restricting geometric Fourier analysis of the Lie group
$\mathbf{SL(}2,\mathbb{C)}$ to the image plane of the conformal camera. Thus,
the conformal camera comes with its own harmonic analysis. Moreover, PFT is
computable by FFT in log-polar coordinates given by a complex logarithm that
approximates the retinotopy. It implies that PFT can integrate the head, eyes,
and visual cortex into a single computational system. This aspect is
discussed, with special attention to perisaccadic perception, in the remaining
part of the paper. Another advantage of PFT is the complex (conformal)
geometric analysis underlying the conformal camera. We discussed, in Section
4.1, the relation of this geometry to the intermediate-level vision problem of
grouping local contours into individual objects and the background of natural scenes.

The other approaches to space-variant vision use the geometric
transformations, mainly based on a complex logarithmic function between the
nonuniform (retinal) sampling grid and the uniform (cortical) grid for the
purpose of developing computer programs. These approaches can be classified
into two different groups. The first group of problems deal with visualizing
and classifying large information data sets. We give two examples for the
first group. The first deals with the problem of mapping information space to
the image space for navigation through complex two-dimensional data sets when
viewing small details and at the same time the general overview \cite{bottger}%
. The second gives the model based image processing in mathematical morphology
for qualifying/segmenting/quantifying spots topology in genomic
microarray-based data \cite{angulo}. The second group of problems is related
to robotic vision. We give only a few examples of such problems, which include
tracking \cite{bernardino1}, navigation \cite{baratoff}, detection salient
regions \cite{tamayo}, and disparity estimation \cite{manzotti}. However, it
seems that they share one common problem: high computational costs in the
geometric transformation process.

In the next figure, we show a simulation applied to Fig. 2 (a) with the
software available over internet \cite{bernardino0}. In Fig. 6, the San Diego
skyline and harbor shown in (a) is sampled in retinal exp-polar coordinates
(with the vertical meridian deleted according to the split theory discussed
before) and mapped by a complex logarithm transformation to rectangular
log-polar coordinates (b). The inverse geometric transformation shown in (c)
results in the retinal image that simulates the sampling by the ganglion cells
density as a function of eccentricity.%
\begin{figure}
[h]
\begin{center}
\includegraphics[
natheight=1.649800in,
natwidth=4.402800in,
height=1.6498in,
width=4.4028in
]%
{/document/graphics/6__6.jpg}%
\caption{(a) San Diego skyline and harbor. (b) Its log-polar image, the
vertical meridian deleted, obtained by the geometric transformation of both
the polar samples with the radial partition changing exponentially and a
constant angular partition, into regular samples in log-polar rectangular
plane (c). }%
\end{center}
\end{figure}

We note that the image processing presented here (see the last paragraph in
the previous section) differs from the above simulation by one crucial aspect:
we use projective Fourier analysis framework for image representation that
provides low computational cost of the retino-cortical (logarithmic) transformation.

\subsection{DPFT and Binocular Vision}

In order to carry out numerical experiments with the discrete PFT, the
conformal camera should work in the following setup: we get a set of samples
$f_{k,l}=f(e^{u_{k}}e^{i\theta_{l}})$ of an image $f$ from a camera with
anthropomorphic visual sensors \cite{berton} or an `exp-polar' scanner with
the sampling geometry similar to the distribution density of the retinal
ganglion cells. Next, we form DPFT $\widehat{f}_{k,l}$ according to
(\ref{lpdft}) and compute it with FFT. Then, we compute IDPFT of $\widehat
{f}_{k,l}$ given in (\ref{ilpdft}), again with FFT. However this output from
IDPFT renders the retinotopic image $\mathrm{f}_{k,l}$ of the retinal samples
in cortical log-polar coordinates. This setup provides an efficient model that
integrates the head, eyes, and the cortex into a single computational system,
which is introduced next.

We discuss this integrated system by assuming that a $3$D scene consists of a
gray square with a red bar located in front of it (see Fig. 7).%
\begin{figure}
[h]
\begin{center}
\includegraphics[
natheight=1.720100in,
natwidth=3.721300in,
height=1.7201in,
width=3.7213in
]%
{/document/graphics/7b__7.jpg}%
\caption{The scene consisting of a gray square with a red bar in front of it
is seen by an observer. The visual pathway with the major cortical areas is
shown.}%
\end{center}
\end{figure}
The integrated binocular system with eyes modeled by the conformal cameras and
this scene as seen from above is shown in Fig. 8.%
\begin{figure}
[hh]
\begin{center}
\includegraphics[
natheight=2.287200in,
natwidth=4.193600in,
height=2.2872in,
width=4.1936in
]%
{/document/graphics/8b__8.jpg}%
\caption{The head-eyes-visual cortex integrated system. Following from the
fact that eyes are modeled by the conformal camera, theoretical horopters are
conics that resemble empirical horopters.}%
\end{center}
\end{figure}

A simulation of the integrated binocular system with the grey square-red bar
scene can be seen in Fig. 9. Each eye sees the scene from a different vantage
point ((a) and (c) in Fig. 9), as the eyes are separated laterally. The
retinal projections are sampled on the exp-polar grid with the meridian line
removed as implied by the split theory.%
\begin{figure}
[h]
\begin{center}
\includegraphics[
natheight=2.805300in,
natwidth=4.340700in,
height=2.8053in,
width=4.3407in
]%
{/document/graphics/9b__9.jpg}%
\caption{In (a) and (c), the $3$D scene from Fig. 8 is seen from a different
vantage point by each eye (i.e., the conformal camera) due to eyes lateral
displacement. The Matlab-simulated right and left retinal projections and the
retinotopic image can be seen in (b), (d) and (f), respectively.}%
\end{center}
\end{figure}

The retinotopic images are simulated in Matlab using the program from
\cite{bernardino0}, and the cut-and-paste transformations are used to account
for the global retinotopy topology. For example, the output from FFT computing
the inverse DPFT of the scene projected on the right eye and sampled by
ganglion cells is shown in Fig 10 (b).%
\begin{figure}
[h]
\begin{center}
\includegraphics[
natheight=1.171800in,
natwidth=2.147500in,
height=1.1718in,
width=2.1475in
]%
{/document/graphics/10__10.jpg}%
\caption{(a) The simulation of the rigth eye's projected scene sampled by
ganglion cells. (b) The retinotopic image of the sampled projection shown in
(a); the vertical size corresponds to the lenght of the angular interval
$\left[  -\pi,\pi\right]  $ .}%
\end{center}
\end{figure}
The cut-and-paste operation is applied to the output in Fig. 10 (b) and to the
corresponding DPFT output of the left eye to obtain (f) in Fig. 9.

\subsection{Modeling Perisaccadic Perception with DPFT}

Because of acuity limitations of foveate vision, a sequence of fast eye
rotations is necessary for processing the details of the scene by fixating
eyes consecutively on the targets of interest. The sequence of fixations,
saccades and smooth pursuits, called the scanpath, is the most basic feature
of foveate vision (cf., Fig 2). The fact that we do not see moving images on
the retinas points to a poor integration of visual information across
fixations during tasks such as reading, visual searching, or looking at a
scene. Given the limited computational resources, it is critical that visual
information is not only efficiently acquired during each fixation, but also
that it is done without starting anew much of this acquisition process at each fixation.

Although we are not aware of discontinuities in a scene perception when
executing a scanpath, this visual constancy is not perfect. In psychophysical
laboratory experiments, the phenomenon of perisaccadic compression is
observed: before the onset of the saccade, brief flashes are perceived by
human subjects to be compressed around the impending saccade target
\cite{Kaiser,Ross}, see Fig. 11. However, perisaccadic perception experiments
have revealed a multitude of mislocalization phenomena, pointing to the
involvement of many different neural processes. Accordingly, many different
theories have been proposed, see \cite{Hamker2}.%
\begin{figure}
[h]
\begin{center}
\includegraphics[
natheight=2.642600in,
natwidth=3.981100in,
height=2.6426in,
width=3.9811in
]%
{/document/graphics/11b__11.jpg}%
\caption{The spatial pattern of perisaccadis compression. It shows
experimental data of the absolute mislocalization (lower row), reference to
the true position of flashed dot randomly chosen from an array of 24 dots and
four different saccade amplitude (upper row). Adapted from \cite{Lappe}.}%
\end{center}
\end{figure}

Two computational theories of perisaccadic vision that have been proposed in
visual neuroscience are related to our modeling. The first theory, suggested
in \cite{VanRullen}, states that an efference copy generated by SC, a copy of
an oculomotor command to rotate eyes in order to execute the saccade, is used
to uniformly shift cortical neural activity representing spatial locations of
the saccade target area toward foveal representation. It was proposed that
this shift is reflected on the neuronal level by a transient spatial remapping
of the receptive fields in numerous retinotopically organized cortical areas
(\cite{Duhamel,Krekelberg}), including the superior colliculus (SC), parietal
eye field (PEF), and frontal eye field (FEF). It can explain the perceived
increase in spatial resolution around the saccade target as more foveal
neurons are available there to process the details of objects. Furthermore,
because the shift occurs in logarithmic coordinates that approximate
retinotopy, the model can also explain perceived perisaccadic compression.

The second theory, \cite{Lappe}, explains the perisaccadic compression by
directing spatial attention to the target of a planned saccade. The proposed
computational model assumes that the initial stimulus neuronal activity in the
visual cortical area\ is distorted by the feedback of the retinotopically
organized activity hill of the saccade target in the oculomotor SC layer, what
pushes the population response of the flashed stimulus in retinotopically
organized cortical areas (including PEF and FEF) towards the saccade target.
This boost of performance at the target location of the saccade occurs
immediately before the saccade onset increases spatial discrimination. The
shift of the neuronal activity in logarithmic\ coordinates, and hence
perisaccadic compression, is a direct consequences of it.

Because circuitry underlying receptive field remapping is widespread and not
well understood, it cannot be easily decided whether saccadic remapping is the
cause or consequence of saccadic compression. For example, only recently it
was reported in \cite{Merriam} that a phenomenon very similar to the remapping
occurs in extrastriate (V4, and, though progressively weaker, V3, V2 and V1)
cortical areas in humans. Remarkably, remapping in extrastriate cortex could
be functionally related to the integration of visual information from a
constant object across saccades \cite{Gottlieb}.

In this section, we model perisaccadic perception using the integrated
binocular system, addressing the process of presaccadic activity consisting
shifts of neurons current receptive fields to their future postsaccadic
locations, that is thought to underlie the scene remapping based on
anticipated saccadic eye movement (efference copy) with the accompanied
perisaccadic perceptual space compression. The postsaccadic activity during
which actual integration of visual features takes place, will be considered in
the next stage of our modeling. Although, our modeling directly conforms to
the theory in \cite{VanRullen}, it may also be useful, on the image processing
level, in representing the resulting receptive field shifts from `attentional
multiplicative gain field interaction'\ \cite{Lappe}, especially since the
efficiency of the whole modeling, which must be repeated three times per
second, was not addressed by the authors.

We start here by supplementing the integrated binocular system presented in
Section 6.4 with the most important subcortical and cortical pathways of the
visuo-saccadic neural processes. These pathways depicted by arrowhead lines in
Fig. 12, include the SC of the midbrain, which contains retinotopically
organized visual and oculomotor layers, the PEF, and the FEF in the parietal
and frontal lobes of the neocortex (which themselves obtain inputs from many
visual cortical areas) for assisting the SC in the control of the involuntary
(PEF) and voluntary (FEF) saccades. We also include the interhemispheric
pathways, the corpus callosum (about $500$ million of neuronal axons
connecting cerebral cortical hemispheres), and the intercollicular commissure,
because the coordinated movement of two eyes is a bihemispheric event. The
motor commands that originate from the brain's major hemisphere (the left
hemisphere for most right-handed people) travel across the corpus callosum to
the minor hemisphere then down to brainstem, where part of it again crosses to
the other side of the brain before both eyes are finally moved in coordination
\cite{Derakhshan}.%
\begin{figure}
[h]
\begin{center}
\includegraphics[
natheight=4.786800in,
natwidth=3.803000in,
height=4.7868in,
width=3.803in
]%
{/document/graphics/12b__12.jpg}%
\caption{The description is given in the text of the article.}%
\end{center}
\end{figure}
We believe that building on the existing models \cite{Girard} and accelerating
advances in visual neuroscience will soon allow the inclusion of these
pathways such that the operation of a more complete system of perisaccadic
perception can be addressed in numerical modeling in a way that could be
useful in neural engineering designs.

The course of events taking place during perisaccadic perception, shown in
Fig. 12, is as follows: the eyes are fixated at \textbf{F }and the new
stimulus appears at \textbf{T. }The SC population \textbf{T}$^{\prime}%
$\textbf{\ }at the retinotopic image of \textbf{T} (green spot in the left SC)
calculates the position of the target \textbf{T} of an impending saccade. The
SC also codes the motor command for the execution of the saccade.

About $50$ ms before the onset of the saccade, during the saccade (about $30$
ms), and about $50$ ms after the saccade, the visual sensitivity is reduced
and flashes (dark blue dotes) around \textbf{T} are not perceived in veridical
locations. Instead, a copy of the motor command (efference copy) is sent to
translate the cortical image (light blue dots in V1) of flashes to remap it
into a target-centered frame (red dotes in V1).

This internal remapping results in the illusory compression of flashes, shown
by red arrows. The compression is perceived around the incoming target
\textbf{T} even though the eyes fixation is moving from \textbf{F} to
\textbf{T}. The location of the cortical area of neural correlates of
remapping is uncertain; it is required that this area is retinotopically
organized. Although it could be PEF/FEF, here, for simplicity, this area is
represented by V1.

During the fixation of eyes at \textbf{F},\textbf{\ }lasting on average about
$300$ ms, the image is sampled by ganglion cells $f_{k,l}=f(e^{u_{k}%
}e^{i\theta_{l}})$ and its DPFT $\widehat{f}_{k,l}$ is computed by FFT in
log-polar coordinates $(u_{k},\theta_{l})$ where $u_{k}=\ln r_{k}$. The
inverse DPFT, computed again by FFT, gives a cortical image representation
\[
\mathrm{f}_{k,l}=\mathrm{f}(u_{k},\theta_{l})=f(e^{u_{k}}e^{i\theta_{l}})
\]
where disparity-sensitive cells contribute to the building $3$D understanding
of the scene. In the same fixation period, the next saccade's target \textbf{T
}is selected (PEF/FEF) and its position in respect to the fovea is calculated
and converted into the motor command to move the eyes (SC). During that time
interval of about $130$ ms, the visual sensitivity is reduced, neural
processes, using a copy of the eyes motor command (efference copy),
transiently shift the cortical image. In our modeling, this shift is generated
using the shift property of Fourier transform as follows%
\[
\mathrm{f}(u_{k}+j\delta,\theta_{l})=\mathrm{f}_{k+j,l}=\frac{1}{MN}%
{\displaystyle\sum\limits_{k=0}^{M-1}}
{\displaystyle\sum\limits_{l=0}^{N-1}}
e^{i2\pi mj/M}\widehat{f}_{k,l}e^{-(u_{k}+j\delta)}e^{i2\pi mk/M}e^{i2\pi
nl/N}%
\]
where $\delta$ is the corresponding spacing. It brings the presaccadic scene
at \textbf{F }in fovea-centered coordinates into postsaccadic scene \textbf{T
}in target-centered coordinates. However,%
\[
\mathrm{f}(u_{k}+j\delta,\theta_{l})=f(e^{u_{k}+j\delta}e^{i\theta_{l}%
})=f(e^{j\delta}r_{k}e^{i\theta_{l}})
\]
compresses perceptual space.

\subsubsection{Challenges with Implementation}

There are some problems that must be addressed before we can implement our
modeling of primate perception. One problem is due to the fact that the model
of retinotopy is fovea-less. The other is related to the global topology of
retinotopy, and in particular, to the vertical meridian split in retinas (and
hence in the visual field) of the brain's hemispheric projections.

In order to address the first problem, we need to develop a model of
retinotopy that will include both foveal and peripheral regions. Hence, the
projective Fourier transform that gives extrafoveal image representation must
be complemented with a transform for the foveal image representation. Two
different transforms, foveal and extrafoveal, could conform to the accumulated
evidence indicating that the fovea and periphery have different functional
roles in vision and may have visual processing differences
\cite{petrov,prado,Xing}. Maybe the simplest way to construct the foveal image
transform is restricting the group $\mathbf{SL(}2,\mathbb{C)}$ action (which
gives both image projective transformations and M\"{o}bius geometry), to
Euclidean or affine subgroups. We refer to Section 3.2 in \cite{Turski4} where
Euclidean Fourier transform is introduced in the framework of representation
theory to motivate the construction of PFT and can be seen as its
`restriction' to the Euclidean subgroup of $\mathbf{SL(}2,\mathbb{C)}$. The
affine subgroup could bring the wavelet transform to supplement PFT.

The second problem involves two facts that are not compatible with each other:
the computation of DPFT of an image in log-polar coordinates by FFT and the
foveal split along the vertical meridian and partial crossing that
re-organizes the retina outputs so that the left hemisphere destinations
receive information from the right visual field, and the right hemisphere
destinations receive information from the left visual field. The retina (that
is, the image plane of the conformal camera) with the foveal disc removed has
the visual field representad by an annulus, which under the complex logarithm
$w=\ln z$ is mapped into a rectangle. In order to discretize PFT, this
rectangle must be extended periodically, which forces a quasiperiodic
extension of the annulus, see Eq. 21 in \cite{Turski1}. In our numerical
experiments with the image translation by the corresponding shift property of
DPFT, the image `disappeared' into the foveal region of the cortical area (the
foveal region of the retina) to reappear from the opposite side of the
rectangle (opposite circular boundary of the annulus). Also, we need to modify
the FFT to account for the global retinotopy simulated in Fig. 9 (f) by the
cut-and-paste transformations. Clearly, the two problems are interdependent.

\section{Conclusions}

In this article we presented a comprehensive account of our approach to
computational vision developed over the last decade. It was done by bringing
in one place physiological and behavioral aspects of primate visual perception
and the conformal camera's computational harmonic analysis with the underlying
geometry. This allowed us to discuss remarkable advantages that the conformal
camera possesses over other cameras used in computational vision. First, the
conformal camera geometry fully accounts for the basic concepts of
co-circularity and scale invariance employed by human vision system in solving
the difficult intermediate-level vision problems of grouping local elements
into individual objects of natural scenes. Second, the conformal camera has
its own harmonic analysis---projective Fourier analysis--- for image
representation and processing that is well adapted to image projective
transformations and the retinotopic mapping of the brain visual and oculomotor
pathways. Projective Fourier analysis integrates the binocular model
consisting of the head, eyes (conformal cameras), and the visual cortex into a
single computational system. Based on this binocular system, we proposed a
model of the perisaccadic perception, including perisaccadic mislocalizations
observed in laboratory experiments. More precisely, we modeled the presaccadic
activity, which, through shifts of neurons current receptive fields to their
future postsaccadic locations, is thought to underlie remapping based on
anticipated saccadic eye movement (efference copy). The postsaccadic activity,
during which the actual integration of visual features takes place, will be
considered in the next stage of our modeling.

Finally, we presented numerous challenges with the implementation of our
modeling. First, the fovea-less model of the retina, based on the discrete
projective Fourier transform (DPFT) of an image, must be supplemented with the
foveal image transform. Second, the computations of the DPFT with a fast
Fourier transform algorithm (FFT) has to be modified in order to account for
the global retinotopy of the brain visual pathway.

It was observed that saccades cause, not only a compression of space, but also
of time \cite{Morrone}. In order to preserve visual stability during the
saccadic scanpath, receptive fields undergo a fast remapping at the time of
saccades. When the speed of this remapping approaches the physical limit of
neural information transfer, relativistic-like effects are
psychophysiologically observed and may cause space-time compression
\cite{Burr,Morrone2}. Curiously, this suggestion can also be accounted for in
our model based on projective Fourier analysis since the group of image
projective transformations in the conformal camera is the double cover of the
group of Lorentz transformations of Einstein's special relativity.

\end{document}